# Tackling problems of marker-based augmented reality under water

Authors: Jan Čejka, Fotis Liarokapis

## Abstract

Underwater sites are a harsh environment for augmented reality applications. Obstacles that must be battled include poor visibility conditions, difficult navigation, and hard manipulation with devices under water. This chapter focuses on the problem of localizing a device under water using markers. It discusses various filters that enhance and improve images recorded under water, and their impact on marker-based tracking. It presents various combinations of 10 image improving algorithms and 4 marker detecting algorithms, and tests their performance in real situations. All solutions are designed to run real-time on mobile devices to provide a solid basis for augmented reality. Usability of this solution is evaluated on locations in Mediterranean Sea. It is shown that image improving algorithms with carefully chosen parameters can reduce the problems with visibility under water and improve the detection of markers. The best results are obtained with marker detecting algorithms that are specifically designed for underwater environments.

## Introduction

Cultural heritage sites and artefacts are spread all around the word, and people search them to learn more about their history and lives. Today, they are not limited only to observe these objects in their current state and read about their story, but thanks to modern technologies like augmented reality (AR), they can see these objects as virtual models superimposed into the real world to see how they fit into the scene and how they interact with other objects. These technologies are able to show even missing parts of settlements or whole buildings [1, 2].

Historical artefacts are not only on land, but many of them are hidden under water. This includes wrecks of ancient ships transporting goods between cities, or seaside settlements that submerged over the last thousands of years. Unfortunate-ly, underwater sites are not only harder to access for people that wish see the artefacts, but they also impose many problems for technology to work. Localization techniques based on GPS, Wi-Fi, or Bluetooth technology do not work as their signal is absorbed very quickly. Augmented reality and other computer vision solutions that require visual input struggle with problems like low contrast of images, sensor noise caused by recording images in low light, occlusions caused by small particles and fish floating in water, unnatural colors due to uneven absorption of light, and short visibility limited due to turbidity.

The idea of using AR under water is not new [3], however, solutions are limited mostly on the clear water of swimming pools [4, 5, 6]. In marine areas, AR solutions use acoustic beacons to replace the visual input [7, 8], but they are limited only to show a map and a textual information about the area, since they are not able to track precise position of the diver required to accurately superimpose virtual objects. Impact of bad visibility conditions on algorithms of computer vision was tested in laboratory conditions [9] or as a part of an evaluation of a single image improving algorithm [10, 11], but such evaluation was not focused on sea environments.

This chapter describes solutions used for improving marker-based tracking for project iMareCULTURE [12] and is based on results of Žuži et al. [13] and Čejka et al. [14, 15]. These works focus on postprocessing images taken under water to increase their quality to improve detection of markers for AR. It is divided into three parts. First, it evaluates nine algorithms for improving images to assess their performance for enhancing a quality of images before detecting markers for AR. Second, it chooses the most promising solutions and performs a deeper analysis in various sea environments. The final part inspects which components of marker detecting algorithms are affected in in bad visibility conditions and presents a results of a cultural heritage use case scenario.

## Performance of image-enhancing algorithms

Visibility under water is affected by many factors that are hard to separate to explore their impact individually, so for this reason, all algorithms in this chapter are evaluated using a data recorded in sea environments. First, to get an initial insight about the performance of image-enhancing algorithms, this section presents a brief analysis of nine solutions by recording a video with markers placed under water, processing it offline, and comparing the number of detected markers. The most promising solutions are chosen and studied in more details in the following sections.

The test was conducted with a video with two divers holding a sheet with markers (see Figure 1). A camera operated by another diver is moving closer to them to decrease the impact of bad under water visibility conditions when the distance to the markers gets smaller. The video is recorded with a GoPro camera with resolution of 1920 × 1080, in the Mediterranean Sea near Athens in depths ranging from 5 to 7 meters. At the beginning of the video, the size of the smaller markers is roughly 20 pixels to assess the influence of turbidity on detection of small distant markers. The larger marker is roughly 85 pixels in the beginning of the video, which focuses more on the problem of turbidity if the size of the marker is just a minor issue.

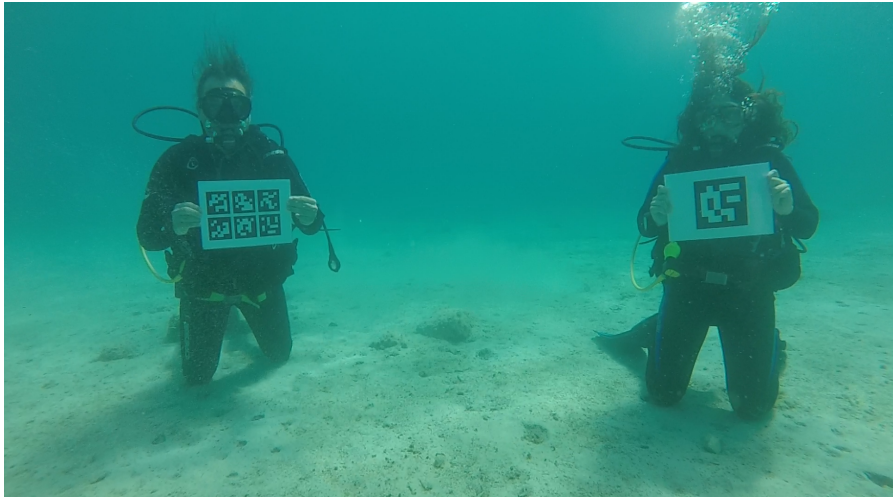

*Figure 1: Two divers holding sheets with markers for initial assessment of image-improving algorithms.*

ARUco library [16] is used to detect markers in images. This library is open-source, kept up-to-date, robust to different lighting conditions, runs in real time, and as shown in [9], it provides good results in a reasonable time. Its implementation is a part of OpenCV 3.2.0. It detects markers in gray-scale images, so prior the detection, all images are converted to YUV color space. This color space was chosen, because it is natively supported by many mobile devices. The color conversion is done before or after enhancing original images, depending on the enhancing algorithm.

The sheets contained seven markers. To identify individual markers, ARUco uses a binary matrix of 6 rows and 6 columns to create a code with 36 bits. Thanks to this, it can correct up to 6 incorrectly detected bits. Six markers were printed on a A4 paper, formed in a grid of two rows and three columns. Each marker measured approximately 7 centimeters with 1 centimeter of white space between them. The seventh marker was printed larger with size of approximately 15 centimeters on a separate paper. This was decided in order to evaluate the performance of two potential settings: single marker tracking (in which a one marker superimposes one object), and multi marker tracking (where multiple markers superimpose one object). Regardless, the detection of each marker of this multi marker was evaluated separately, to obtain finer results. It is worth mentioning that the markers were plasticized in order to 'survive' in underwater environments.

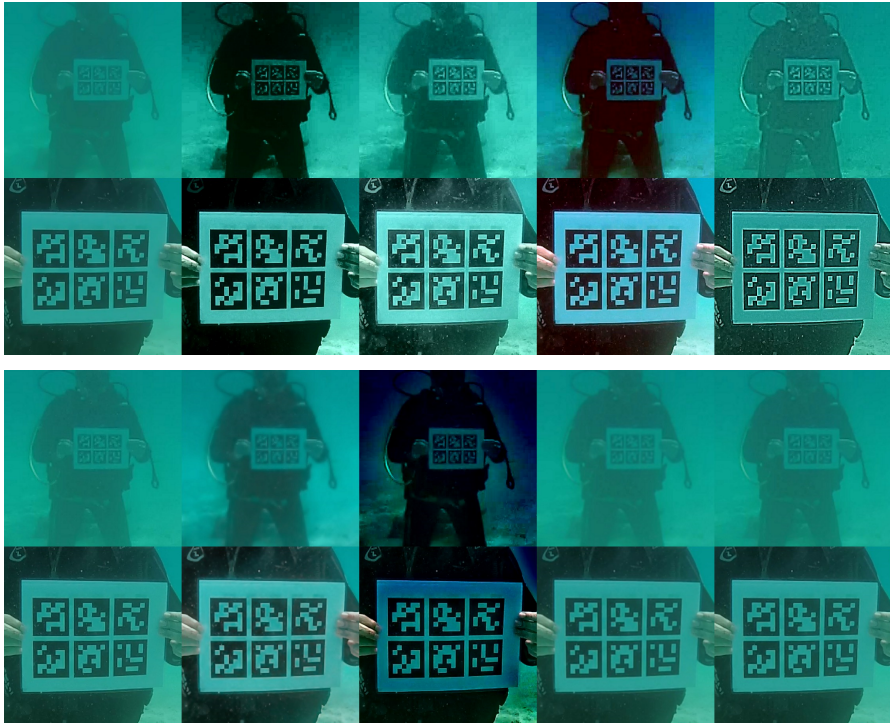

*Figure 2: Image enhancing algorithms. Top row from left to right: original image, fusion, bright channel prior, Gaussian filter, median filter; bottom row from left to right: bilateral filter, histogram equalization, contrast limited adaptive histogram equalization, white balancing, debluring.*

The video was processed off-line on a standard PC with processor Intel Core i5 760, 8 GB of operating memory, and operating system Windows 10. Each frame was processed separately, first increasing its quality by applying image-improving algorithms, and then detecting markers.

**Tested image enhancing algorithms**

In total, nine image-improving algorithms were evaluated: two algorithms designed to improve the colors of underwater images (fusion based on [17], and bright channel prior based on [11]), three algorithms for denoising (Gaussian filter, median filter, and bilateral filter), three algorithms for increasing contrast (histogram equalization, contrast limited adaptive histogram equalization, and white balancing), and one algorithm for image sharpening (debluring), see Figure 2. The algorithms were implemented in MATLAB, OpenCV 3.2.0, and C++. To assess them, different parameters were explored to find out how they affect tracking accuracy. Algorithms based on neural networks [18, 19] were not considered in this experiment, as they require a large amount of data for training.

*Fusion*

The fusion algorithm is based on [17]. It restores the colors and enhances the contrast of images taken under water by deriving two improved versions of the input image and combines them together using a multi-scale fusion process with a weight function computed from each image.

The restoration process is simplified by using general image-improving techniques, instead of choosing a more sophisticated technique based on the physical properties of the scene. The first enhanced version of the input image is obtained by using a white balancing algorithm that corrects the shift of colors caused by various illuminations of the scene. This algorithm is based on the Gray-World approach [20], which was found as the most appropriate method for underwater images. The second enhanced version of the input image is aiming at reducing noise and improving contrast of the underwater image, by applying the bilateral filter on the color-corrected image created in previous step. The contrast is further improved by applying a local adaptive histogram equalization method.

Fusion of the derived images is controlled by local contrast weights derived from the luminance of the images, and saliency weight to highlight the objects with higher importance. The final image is created by blending the two derived input images by their normalized weight values. The implementation was done in Matlab, where most of the filters and operations are available directly, using its Image Processing Toolbox.

*Bright channel prior*

Bright channel prior (BCP) is a technique developed by [11], which is based on a method of [21] to dehaze images taken on land that are affected by fog. [21] observes that with images taken on land with no fog, intensity of one color channel is very low. However, this assumption fails to work for images taken under water, since the colors are shifted due to an uneven absorption of light under water. Visible light with longer wavelength is absorbed more quickly, which results in underwater scenes being green or blue due to very low intensities of red color channel. Because of this, [11] define the bright channel image as follows:

$$J^{bcp} = \max_{y \in \Omega(x)} \left( \max_{c \in \{r,g,b\}} I^{new}(y) \right)$$

where $\Omega(x)$ denotes a window neighborhood centered at pixel $x$ and $I^{new}$ is the input image with original red color channel and inverted green and blue channels. Intensities of all color channels of this bright channel image are close to one for pixels without haze, and lower for pixels affected by haze and turbidity.

To estimate atmospheric light, which is defines the color distortion of the image, [11] takes one percent of darkest pixels, and a pixel with the least variance is selected as the color of atmospheric light. Transmittance is then derived from the bright channel image and estimated atmospheric light by following equation

$$t^c(x) = \frac{(J^{bcp}(x) - A^c)}{1 - A^c}$$

where $c$ denotes red, green, and blue color channels, $t^{c(x)}$ denotes the transmittance of this channel, and $A^c$ denotes the value of atmospheric light. The final transmittance is obtained by averaging values across the all color channels, which is then processed with guide filter [22] to remote halos. The final haze-free image is restored by formula:

$$J(x) = \frac{I(x) - A}{t(x)} + A$$

where $J$ is the haze-free image, $I(x)$ is the degraded underwater image, $A$ is the previously estimated atmospheric light and $t(x)$ is the transmittance of the scene. BCP algorithm was implemented in Matlab using Image Processing Toolkit.

*Gaussian filter*

Gaussian filter is a standard filter for reducing noise in images. It is defined as:

$$I^{out}(x) = \frac{1}{W} \sum_{x_i \in \Omega} I^{in}(x_i) G_{space}(|x_i - x|)$$

with normalizing factor:

$$W = \sum_{x_i \in \Omega} G_{space}(|x_i - x|)$$

where $G_{space}$ is a Gaussian function with parameter $\sigma_{space}$. Gaussian filter is a linear filter that sums neighbor pixels using weights based on their mutual distance in image space. It decreases the number of edges found by edge detecting algorithms, and thus improves the image for marker detection. Gaussian filtering was applied on the Y channel of our images after their conversion to YUV color space.

*Median filter*

Median filter is another standard filter for reducing noise in images and can be defined as:

$$I^{out}(x) = median_{x_i \in \Omega}(I^{in}(x_i))$$

Median filter is used for removing a noise in image like Gaussian filter, but unlike it, it provides good results when removing impulse noise. Median filtering was applied on the Y channel after the images are converted to YUV color space.

*Bilateral filter*

Bilateral filter is a variation of Gaussian filter, which preserves edges while simultaneously removing noise in smooth areas. It is defined as:

$$I^{out}(x) = \frac{1}{W} \sum_{x_i \in \Omega} I^{in}(x_i) G_{space}(|x_i - x|) G_{color}(|I(x_i) - I(x)|)$$

with normalizing factor:

$$W = \sum_{x_i \in \Omega} G_{space}(|x_i - x|) G_{color}(|I(x_i) - I(x)|)$$

where $G_{space}$ is a Gaussian function with parameter $\sigma_{space}$, and $G_{color}$ is a Gaussian function with parameter $\sigma_{color}$. As Gaussian filter, it sums the weighted intensities of neighbor pixels, but unlike it, the weights do not depend only on the spacial distance of the weighted pixels in image space, but also on the different between intensities of these pixels in color space, allowing the filter to decrease the amount of blurring over edges. The implementation of bilateral filter is based on [23] and is applied to Y channel after the input image is converted to YUV color space.

*Histogram equalization*

Histogram equalization technique is a method that maps intensities of pixels of the original image to different values to balance the histogram of the filtered image. This test uses an ordinary histogram equalization implementation from OpenCV. It was applied on the Y channel after the images were converted to YUV color space.

*Contrast limited adaptive histogram equalization*

Contrast limited adaptive histogram equalization (CLAHE) [24] is an adaptation of histogram equalization, which works with a histogram of a small window around each pixel and reduces the contrast of output image by clipping the highest values of the histogram. CLAHE is applied to Y channel after the images are converted to the YUV color space. CLAHE has a single parameter (clip limit), which influences the amount of values clipped in the histogram.

*White balancing*

White balancing algorithm changes the colors of input image to render white objects correctly under different illuminations like sun or clouded sky. In this test, we used an algorithm presented in [25]:
    **foreach** *color channel* **do**
      compute histogram of this channel;
      $channel_{min} \leftarrow$ *black*-th percentile of values in histogram;
      $channel_{max} \leftarrow$ *white*-th percentile of values in histogram;
      linearly transform all intensities so that $channel_{min} = 0$ and $channel_{max} = 255$;
    **end**
The main advantages of this method include speed and simplicity. Though the restored image may not represent the colors of objects properly, due to the simplicity of the algorithm, this is not a problem, since the image is not presented to a viewer, but it is only processed by a marker detection algorithm. The algorithm was applied to all channels of RGB image before it is converted to YUV space for marker detection.

*Debluring*

The last filter used is a debluring filter (or unsharp mask filter [26]), which emphasizes high frequencies in input image by subtracting its low frequencies from itself as shown below:

$$I_{out} = (1 + w) \cdot I_{in} - w \cdot Gaussian(I_{in}, \sigma_{space})$$

where $w$ represents the weight of subtracted low frequencies, and $Gaussian(I_{in}, \sigma_{space})$ is a Gaussian filter with standard deviation $\sigma_{space}$ applied to the input image $I_{in}$. This method was implemented and applied on the Y channel after the image was transformed into the YUV color space.

## Results of enhancing underwater images

We experimented with different values of parameters for each technique and counted the number of newly found markers that were detected in the enhanced video and not detected in the original video, and a number of lost markers that were detected in the original video and not in the enhanced video. All image enhancing algorithms were also compared between each other, counting the number of detected markers in images processed with one algorithm and not detected in images processed with the other algorithm. To reduce the number results to present, the following tables show the results of only the best parameters for each algorithm that obtain the highest number of newly detected markers while keeping the number of lost markers as low as possible. The following parameters were selected:

- Gaussian filter, $\sigma_{space}$ = 0.4;
- Median filter, window size = 3;
- Bilateral filter, $\sigma_{color}$ = 2.0 and $\sigma_{space}$ = 4.0;
- CLAHE, clip limit 2;
- White balancing, *black / white* percentile 2/98;
- Debluring, $\sigma_{space}$ = 2.8 and weight = 1.9.

The evaluation of the results is shown in Table 1 and demonstrates that the white balancing algorithm provides the best outcome, followed by debluring, then CLAHE, and BCP. Other algorithms provided similar or worse results then with the original image. Also, it is worth noting that Gaussian filter parameter $\sigma_{space}$ is very low, which indicates that high amount of blur worsens the results.

Performance of the algorithms depend on the chosen parameters. Improper parameters for CLAHE and white balancing algorithms make the results of detection worse than when detecting markers in the original images. It should be also noted that the best results are obtained when using very low values of parameters, e.g., in case of debluring, sufficiently good results are available even with smaller values.

This test was not focused on measuring the time to process the images, however, we see that real-time algorithms CLAHE, white balancing, and debluring out-performed more sophisticated BCP. This shows that even a simple real-time algorithm can improve the detection of markers under water.

| Algorithm | Found Markers | Number of found markers that were not found by following algorithms | | | | | | | | | |
|---|---|---|---|---|---|---|---|---|---|---|---|
| | | Original image | Fusion | BCP | Gaussian filter | Median filter | Bilateral filter | Hist. equalization | CLAHE | White balancing | Debluring |
| Original image | 4811 | 0 | 231 | 35 | 57 | 312 | 30 | 625 | 43 | 28 | 20 |
| Fusion | 4611 | 31 | 0 | 16 | 26 | 152 | 27 | 457 | 17 | 9 | 8 |
| BCP | 5097 | 321 | 502 | 0 | 336 | 592 | 309 | 876 | 91 | 83 | 71 |
| Gaussian filter | 4795 | 41 | 210 | 34 | 0 | 298 | 48 | 612 | 40 | 24 | 19 |
| Median filter | 4515 | 16 | 56 | 10 | 18 | 0 | 13 | 395 | 13 | 7 | 7 |
| Bilateral filter | 4818 | 37 | 234 | 30 | 71 | 316 | 0 | 628 | 35 | 29 | 20 |
| Histogram equalization | 4238 | 52 | 84 | 17 | 55 | 118 | 48 | 0 | 20 | 14 | 12 |
| CLAHE | 5120 | 352 | 526 | 114 | 365 | 618 | 337 | 902 | 0 | 75 | 58 |
| White balancing | 5264 | 481 | 662 | 250 | 493 | 756 | 475 | 1040 | 219 | 0 | 116 |
| Debluring | 5249 | 458 | 646 | 223 | 473 | 741 | 451 | 1023 | 187 | 101 | 0 |

*Table 1: Results of various image improving algorithms*

**Summary of algorithms enhancing underwater images**

We see that there is no algorithm that would be strictly better than other algorithms. Many algorithms were found to improve the detection of markers when, but in all cases, some markers from the original image were lost. Also, sophisticated offline solutions are not necessarily better than real-time general solutions. This test shows that the most promising algorithms are algorithms that improve sharpness of the image (debluring), colors of the image (white balancing), and its contrast (CLAHE). In the following section, these algorithms are tested into more depth with more videos and more marker detecting algorithms.

## Methodology of the second test

The second test focuses on the image-enhancing algorithms that gave the best results in the first test and analyses them in additional underwater environments. Also, it adds the results of an improved version of which balancing algorithm, marker-based underwater white balancing (MBUWWB) [14] that is better adapted to underwater conditions. Additionally, it compares the results with AprilTag [27], a marker detecting algorithm that provides better results than ARUco at a higher detection time, as shown in [9].

**Testing sites**

Three additional sets of videos were tested, to the total number of four, including the video from the previous experiment, see Table 2.

The first set of videos is the video tested in the previous section. As already noted, it was taken in the depth of approximately 7 to 9 meters in a moderately turbid environment with a GoPro camera with resolution of 1920 × 1080. We refer to this set as Environment 1. The video was recorded using MPEG-4 compression and decoded into RGB frames. The camera starts far from the markers and moves slowly towards the markers.

The second set of videos consists of videos taken in depth of 5 to 6 meters in a highly turbid environment using iPad Pro 9.7-inch with resolution of 1920 × 1080. We refer to this set as Environment 2. These videos were recorded using MPEG-2 compression and decoded into RGB frames. In these videos, the position of the camera from the markers changes from very large distances, where the markers are not distinguishable due to the turbidity, to distances of tens of centimeters.

The third set of videos consists of ten videos taken in the depth of 20 to 22 meters in a moderately turbid environment with a GARMIN VIRB XE camera with resolution of 1920 × 1440. We refer to this set as Environment 3. These videos were recorded using MPEG-4 compression, and decoded into RGB frames. This set of videos contains not only videos with the camera moving towards the markers, but also videos with markers recorded from multiple directions and distances.

| 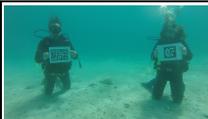 | 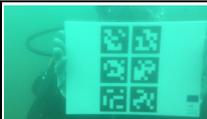 | 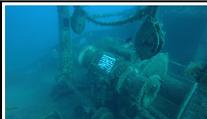 | 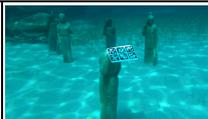 |
|---|---|---|---|
| Environment 1 | Environment 2 | Environment 3 | Environment 4 |
| Moderate turbidity | High turbidity | Moderate turbidity | Low turbidity |
| Depth 5 – 7 meters | Depth 5 – 6 meters | Depth 20–22 meters | Depth 7 – 9 meters |
| GoPro camera | iPad Pro 9.7 | GARMIN VIRB XE | NVIDIA SHIELD |
| 1920 × 1080 | 1920 × 1080 | 1920 × 1440 | 1920 × 1080 |
| MPEG-4 | MPEG-2 | MPEG-4 | MPEG-4 |
| 29.97 fps, 31 sec. | 30 fps, 85 sec. | 24 fps, 160 sec. | 30 fps, 81 sec. |

*Table 2: Four different testing sites*

The fourth set of videos consists of eight videos taken in the depth of 7 to 9 meters in a moderately turbid environment with a NVIDIA SHIELD tablet. We refer to this set as Environment 4. Two of these videos were recorded with resolution of 1920 × 1080 using MPEG-4 compression and decoded into RGB frames. The rest was recorded with resolution of 1280 × 720 and stored without using any compression as YUV frames. As with the third set of videos, this set of videos also contains not only videos with the camera moving towards the markers, but also videos with markers recorded from multiple directions and distances.

**Tested algorithms**

In addition to the best algorithms tested in the previous test, CLAHE, deblurring, and white balancing, this section contains also results of a fourth algorithm, marker-based underwater white balancing (MBUWWB), that is adapted for marker-based tracking under water [14]. This algorithm solves an intrinsic part of white balancing algorithms, which is finding the colors that are subsequently mapped to the white and black in the filtered image. White balancing algorithm define in the previous section chooses these colors as a percentile of values in input image histogram, but MBUWWB assumes that the marker is black and white, and instead of computing the histogram of the whole image, it computes the histogram only of the part of the image which contains markers.

In this experiment, MBUWWB is applied to all channels of RGB image, and afterwards, the improved is converted into YUV space, similarly as with the simple white balancing. The images stored in YUV format are also converted to RGB before they were processed by MBUWWB.

**Results in various environments**

Image enhancing algorithms and marker detection algorithms were compared in two aspects: the total number of detected markers, and performance in different visibility conditions.

Summed number of all markers found by ARUco in every set of enhanced videos are illustrated in Table 3. They clearly demonstrate that all tested algorithms improve detection of markers in strongly turbid environments. In moderately turbid environments, the visibility is better, so the improvement in detection is not very apparent.

The behavior of CLAHE algorithm is hard to predict. Using CLAHE for enhancing images can lead to an improvement in the detection of markers (Environment 1, Environment 2), but also it is able to decrease the number of detected markers (Environment 3, Environment 4). The result is also highly dependent on the value of clip limit. On the other hand, debluring, white balancing, and MBUWWB provide much stable results. The results of debluring clearly shows

12|  | Environment 1 | Environment 2 | Environment 3 | Environment 4 |
|---|---|---|---|---|
| Original video | 5338 | 512 | 5847 | 9925 |
| CLAHE, clip limit 2 | 5670 | 3904 | 5405 | 8227 |
| CLAHE, clip limit 4 | 5541 | 4916 | 4801 | 7276 |
| CLAHE, clip limit 6 | 5399 | 5010 | 4190 | 7057 |
| Debluring, weight 1 | 5696 | 1852 | 6033 | 10105 |
| Debluring, weight 4 | 5804 | 4622 | 6050 | 10372 |
| White bal., perc. 0/100 | 5557 | 3603 | 6053 | 9988 |
| White bal., perc. 3/97 | 5668 | 5159 | 5950 | 9558 |
| MBUWWB, perc. 0/100 | 5781 | 5842 | 6094 | 10026 |
| MBUWWB, perc. 3/97 | 5787 | 6351 | 6069 | 10085 |

*Table 3: Total number of markers detected with ARUco in tested sets of videos enhanced by tested algorithms.*

that using value 4 for subtraction weight w leads to higher number of detected markers.

The results also illustrate that MBUWWB provided better overall results than original white balancing, although the difference is much lower in moderately turbid environments. Careful choice of percentile affects the performance of both these algorithms, especially in highly turbid environments. This difference is more visible in case of WB algorithm, while in case of MBUWWB, percentile 3/97 provided better or nearly the same results when compared to percentile 0/100.

The results for AprilTag are in Table 4. Unlike ARUco, AprilTag algorithm is much more robust to environments with strong turbidity (a similar result was observed by [9]). The detection provides already very good results in original video, and the improvement obtained by enhancing the image is not very large, if any.

|  | Environment 1 | Environment 2 | Environment 3 | Environment 4 |
|---|---|---|---|---|
| Original video | 5925 | 6884 | 5775 | 9141 |
| CLAHE, clip limit 2 | 5868 | 7032 | 5552 | 8119 |
| CLAHE, clip limit 4 | 5667 | 6679 | 4974 | 6655 |
| CLAHE, clip limit 6 | 5381 | 6186 | 4362 | 6326 |
| Debluring, weight 1 | 5850 | 6652 | 5781 | 9351 |
| Debluring, weight 4 | 5474 | 5542 | 5847 | 9526 |
| White bal., perc. 0/100 | 5922 | 6873 | 5648 | 9091 |
| White bal., perc. 3/97 | 5479 | 6476 | 5597 | 8769 |
| MBUWWB, perc. 0/100 | 5925 | 6914 | 5705 | 9067 |
| MBUWWB, perc. 3/97 | 5887 | 6695 | 5697 | 8925 |

*Table 4: Total number of markers detected with AprilTag in tested sets of videos enhanced by tested algorithms.*

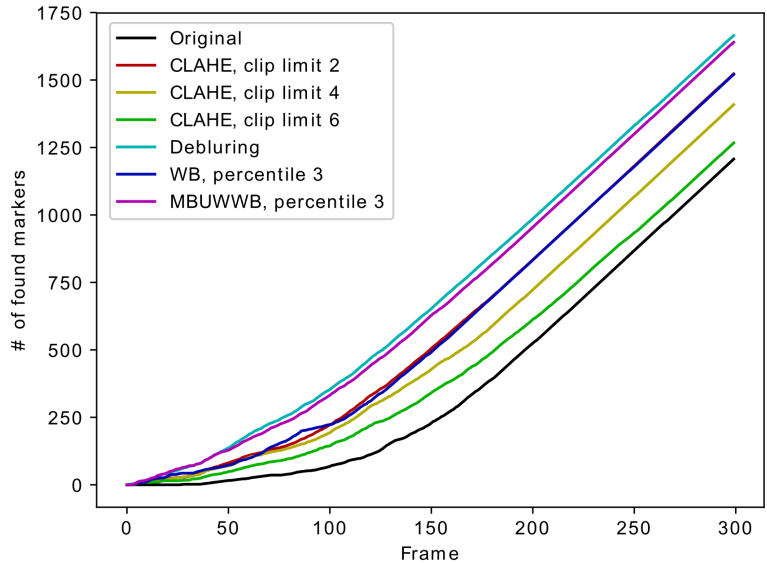

*Figure 3: Number of markers detected by ARUco in a video from Environment 2.*

The best results in individual sets were obtained by CLAHE with clip parameter 2 (Environment 1) and debluring (Environment 3 and Environment 4), although these two algorithms provided much worse results on other videos.

Unlike the results of ARUco, these results show that proper choice of subtraction weight w in debluring algorithm is not easy to get the highest number of detected markers. It can be shown that in environments Environment 1 and Environment 2, using weight 1 lead to a higher number of detected markers than using weight 4, but in environments Environment 3 and Environment 4 the result is the opposite.

The results also show that WB and MBUWWB were not improving the detection, but they are also not making it worse. This is true only when we used percentile 0; the results when using percentile 3 were always worse.

Camera moves towards and away from the markers in some videos of all sets. With decreasing distance, the marker gradually emerges from the turbid, which allows us to evaluate the behavior of image enhancing algorithms in different levels of marker visibility.

We focused on ARUco detector and the video from Environment 1, and plotted the progress of the number of detected markers in time, see Figure 3. In the video, markers start partially in turbid, and as the camera moves towards them, the distance where all the markers are visible is reached very soon.

This experiment shows that in worse visibility conditions (first frames of the video), all image enhancing algorithms improves the detection of the markers, with debluring giving the best results, only slightly better than MBUWWB. Since approximately 150th frame, the visibility is good for all algorithms to find all markers.

### Summary of algorithms performing in various environments

Results indicate the detection of markers can be improved by adding an image improving step, however, proper choice of the algorithm and its parameters is heavily dependent on the environment and the marker-detecting algorithm. If the marker-detecting algorithm is robust, there is no need to improve the input image, its "enhancement" may make the results even worse, but robust algorithms run very slowly. In the final section of this chapter, we will focus on ARUco marker detector to find, which parts are affected by bad visibility conditions.

## Underwater marker-detecting algorithms

The last part of this chapter is focused on the structure of ARUco marker-detecting algorithm to identify parts that are affected by underwater conditions. It also presents results of a cultural heritage use case scenario from Baiae, Italy.

### Structure of marker-detecting algorithm ARUco

The ARUco detector runs fast and reliably recognize markers. First, it thresholds the input image using an adaptive thresholding algorithm, then it finds all contours to detect marker-like shapes, and filters out non-square polygons. After that, it reprojects them to remove perspective distortion, obtains the inner binary marker code, compares it with a dictionary to remove errors, and if correct, it computes the relative position of the marker using its corners.

Čejka et al. in [15] investigates, which parts of ARUco are influenced by image improving algorithms and finds that the most vulnerable part is the initial thresholding. Figure 4 shows that when an underwater image is thresholded by ARUco, the border often breaks, and the marker is not recognized as a square-like object. This can be avoided by changing the parameters of the thresholding, but this step increases the number of small contours, which increases the processing time (a similar method is used in the robust AprilTag algorithm). All four image-improving algorithms tested in previous sections increased the contrast of the image, with approximately the same output with a lesser number of false contours, and so they do not increase the detection time very much.

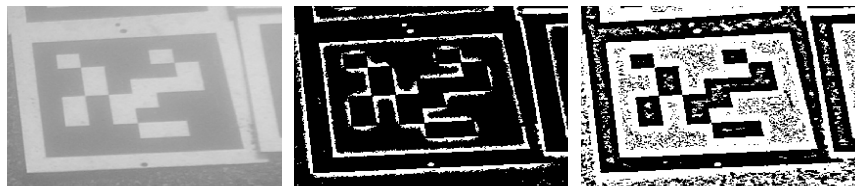

*Figure 4: Left: input image taken under water; Middle: thresholded by ARUco, notice a broken contour; Right: the contour is solid with better parameters of ARUco, but the image contains a lot of noise*

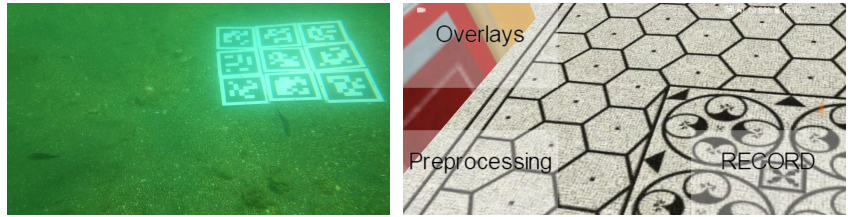

*Figure 5: Left: nine markers placed at the location of a room with mosaic of Villa a Protiro in Baiae, Italy. Right: a virtual model superimposed in augmented reality at the place of these markers.*

|  | ARUco | ARUco with MBUWWB | UWARUco | ARUco3 | AprilTag |
|---|---|---|---|---|---|
| # of markers | 14457 | 19398 | 20145 | 12589 | 20082 |
| Time (ms) | 75.747 | 45.239 | 62.872 | 2.368 | 323.005 |

*Table 5: Number of detected markers and computation time for various augmented reality solutions.*

Čejka et al. in [15] presents an improved version of ARUco called UWARUco that is adapted to underwater environments. This algorithm creates a mask to remove image parts that contain no contour, and applies it to an image thresholded with proper parameters. This approach provides results that are comparable with robust marker detecting algorithms, but their algorithm runs faster.

**Use case: Presenting submerged ancient buildings with AR**

The last part of this chapter is focused on using marker-based augmented reality to present virtual structures to divers that dive in Baiae in Italy at locations of ancient villas. The focus of the testing was limited to one building, Villa a Protiro, with a characteristic mosaic in one of the rooms. Using the improved approach, divers were able to perceive a 3D reconstruction of Villa a Protiro in AR. The test used nine markers from the ARUco DICT_6X6_50 dictionary, forming a grid of 3 × 3 markers. Size of each marker was 19 cm, and the space between markers was approximately 5 cm. The setup and the application are illustrated in Figure 5.

Solutions for augmented reality were tested with a video recorded with Samsung S8 in FullHD resolution. The results are taken from [15] and are presented in Table 5. We see that both solutions that improves the detection of markers, ARUco with MBUWWB and UWARUco, provide better results than original ARUco. ARUco3 [28] is a newer version of ARUco that is optimized for speed and is less robust to underwater environment, which is clearly visible in the results. Robust AprilTag detects high number of markers without the necessity of improving images, but its computation time is very high.

## Conclusion

This chapter described problems with detecting markers in marine underwater environments when targeting applications that use augmented reality to present additional information about cultural heritage sites and superimpose ancient virtual objects. First, it presented a comparison of general solutions that improve images affected by bad visibility conditions under water. It was shown that there is no solution that would perform better than every other solution, and that more elaborated algorithms designed for underwater images do not provide strictly better results than general solutions. Then, the chapter focused deeply on the most promising solutions in multiple underwater environments. Here, the results showed that image improving step can improve the detection of markers if the parameters of the image improving algorithm are correctly set. Finally, it investigated the impact of underwater visibility conditions on a single marker detecting algorithm, and showed that a properly adapted algorithm can outperform general algorithms combined with image enhancing algorithms.

**Acknowledgements**

This research is a part of the i-MareCulture project (Advanced VR, iMmersive Serious Games and Augmented REality as Tools to Raise Awareness and Access to European Underwater CULTURal heritagE, Digital Heritage) that has received funding from the European Union's Horizon 2020 research and innovation program under Grant Agreement No. 727153.